\documentclass{article}
\usepackage[final]{nips_2017} 

\usepackage[utf8]{inputenc} 
\usepackage[T1]{fontenc}    
\usepackage{hyperref}       
\usepackage{url}            
\usepackage{booktabs}       
\usepackage{amsfonts}       
\usepackage{nicefrac}       
\usepackage{microtype}      

\usepackage{times}
\usepackage{helvet}
\usepackage{courier}
\usepackage{caption}
\usepackage{fancyhdr}
\usepackage{time}
\usepackage{graphicx}

\usepackage[ruled]{algorithm2e}

\SetAlFnt{\small}
\SetAlCapFnt{\small}
\SetAlCapNameFnt{\small}
\SetAlCapHSkip{0pt}
\IncMargin{-\parindent}

\title{Faster Fuzzing: Reinitialization with Deep Neural Models}

\author{
  Nicole Nichols, Mark Raugas, Robert Jasper, Nathan Hilliard\\
  Pacific Northwest National Laboratory\\
  1100 Dexter Avenue, Suite 500 \\
  Seattle, WA 98109 \\ ~\\
  \texttt{nicole.nichols@pnnl.gov} \\
 \texttt{mark.raugas@pnnl.gov} \\
 \texttt{robert.jasper@pnnl.gov} \\
  \texttt{nathan.hilliard@pnnl.gov} \\
}

\begin{document}
\maketitle

\begin{abstract}
We improve the performance of the American Fuzzy Lop (AFL) fuzz testing framework by using Generative Adversarial Network (GAN) models to reinitialize he system with novel seed files.
We assess performance based on the temporal rate at which we produce novel and unseen code paths.
We compare this approach to seed file generation from a random draw of bytes observed in the training seed files.
The code path lengths and variations were not sufficiently diverse to fully replace AFL input generation. However, augmenting native AFL with these additional code paths demonstrated improvements over AFL alone.
Specifically, experiments showed the GAN was faster and more effective than the LSTM and out-performed a random augmentation strategy, as measured by the number of unique code paths discovered.
GAN helps AFL discover 14.23\% more code paths than the random strategy in the same amount of CPU time, finds 6.16\% more unique code paths, and finds paths that are on average 13.84\% longer.
Using GAN shows promise as a reinitialization strategy for AFL to help the fuzzer exercise deep paths in software.

\end{abstract}

\section{Introduction}
Identifying software defects through testing is a challenging problem.
Over the years, a number of approaches have been developed to test software, including random mutation testing (black box fuzzing)~\cite{doupe2012enemy,woo2013scheduling}, abstract interpretation (of either source or machine code) ~\cite{cousot1977abstract,cadar2008klee,ma2011directed}, and property based testing ~\cite{arts2006testing,claessen2011quickcheck}.\\

Methods such as symbolic and concolic execution have increased the fidelity of analyses run over programs~\cite{schwartz2010all}.
The development of Satisfiability Modulo Theories (SMT) solvers such as Z3, Boolector, and others have allowed for powerful programmatic analysis of reasoning about software ~\cite{de2008z3,brummayer2009boolector}.
Separation logic has allowed for analyses to be applied to complicated data structures~\cite{reynolds2002separation,dongol2015program}.

American Fuzzy Lop (AFL) is an advanced fuzzing framework that has been used to discover a number of novel software vulnerabilities (\url{https://github.com/mrash/afl-cve}).
AFL uses random mutations of byte strings to identify unique code paths and discover defects in target programs.
The inputs that successfully generated unique code paths are then documented as "seed files".
We propose to use these native seed files as training data for deep generative models to create augmented seed files. 
Our proposed reinitialization methods are a scalable process that can improve the time to discovery of software defects.

Other researchers have used machine learning to augment fuzzing frameworks including: ~\cite{godefroid2017learn}, ~\cite{wang2017skyfire}.
To identify deeper bugs and code paths, Steelix~\cite{Li:2017:SPB:3106237.3106295} uses a program-state based binary fuzzing approach and Driller~\cite{stephens2016driller} demonstrates a hybrid approach using fuzzing and selective concolic execution. AFLFAST~\cite{Bohme:2016:CGF:2976749.2978428} extends AFL using Markov chain models.
Deep Neural Networks (DNNs)~\cite{bengio2015deep}, have had great success in the fields of Natural Language Processing (NLP)~\cite{jones2014learning,wu2016google}, Computer Vision~\cite{krizhevsky2012imagenet}, and the playing of bounded games such as Go~\cite{mnih2013playing} or video games like ATARI~\cite{silver2016mastering}. Can these DNN's help existing program analysis tools perform better?
In our work we investigate that question.
We augment AFL ~\cite{zalewski2015american}, an advanced fuzzing framework, using Generative Adversarial Networks (GAN)~\cite{goodfellow2014generative} and Long Short Term Memory (LSTM)~\cite{sak2014long} to increase its rate of unique code path discovery.

Our work quantifies the benefits that augmentation strategies such as generative models can provide, even when limited by small quantities of training data.
By periodically perturbing the state of AFL as it explores the input space, we are able to improve its performance, as measured by unique code paths.
Specifically, we test our approach on the software ecosystem surrounding the Ethereum ~\cite{wood2014ethereum} project.
As a financial system, correctness of the Ethereum code base is important for guaranteeing that transactions or calculations run without fault.
We choose ethkey as an initial fuzzing target. 
Ethkey is a small C++ program provided as part of the {\tt cpp-ethereum} project used to load, parse, and perform maintenance on Ethereum wallets, and importantly, takes a simple input file, making it easy to test with AFL.

\section{Experimental Design}
First, we describe the basic functionality of AFL, highlighting the key features that connect with the proposed augmentation framework.
Next, we describe the methodology used to create the LSTM and GAN generated seed files.
As a baseline, we also consider random generation of seed files from the training data used to construct the LSTM and GAN models.
AFL has extensions to the GCC compiler which in conjunction with Genetic Algorithms, it uses to create seed files.
Each seed file documents the input that yielded a unique code path, the time of discovery, and 
is used as the basis for mutation (or fuzzing) to generate future seed files.
Our augmentation strategy takes advantage of the fact that if an external tool places additional seed files in the AFL working directory, AFL will use those files as inputs in subsequent fuzzing runs.

To produce the training data for our methods, we run AFL on a target program $P$ for a fixed amount of time.
AFL generates an initial set of seed files $S=\{S_0,...,S_K\}$  for each unique execution trace $\tau$ taken through $P$.
We use $S$ as training examples for the LSTM and GAN models, which are both trained using Keras ~\cite{chollet2017keras}.

Our LSTM is trained from the concatenation of AFL-generated seed file corpus $S$ into a single file and generates new seed files with a maximum length of $40$ characters.
The LSTM model has a $128$ wide initial layer, an internal dense layer, and a final softmax activation layer.
To train the LSTM model, we use RMS propagation as our optimizer and a categorical cross-entropy loss function.
The model takes in a seed sequence sampled from the training corpus and predicts the next character in the sequence.
We additionally tune a separate temperature parameter to diversify the output seed files from the network.
The generated seed files are noted as $S_L$.

In our GAN architecture, two models are built, a generator G, which is pitted against a discriminator D.
G is optimized to generate realistic output, and the discriminator D has the task of predicting if the data is true or fake.
This training strategy is unsupervised and particularly expressive.
The generative model G, is a fully connected 2 layer DNN with a ReLU non-linearity as the inner activation and a tanh output activation. It is trained with a binary cross-entropy loss function via stochastic gradient descent.
The discriminative model D is a 3 layer DNN, but the first layer has 25\% dropout followed by two fully connected layers. It uses an Adam optimizer for stochastic gradient descent and the seed files resulting from the GAN process are noted as $S_G$.

Additionally, given the native AFL seed files, $S$, we randomly draw bytes from this training set and produce new, random seed files $S_R$ of the same length as $S_G$.
This serves as a baseline to determine if the added time and complexity of GAN and LSTM based seed generation are truly providing an advantage over a simple strategy of randomly perturbing AFL's state.


{\bf Small Experiment:}  The seed files ($S_R$, $S_G$, and $S_L$) alone are not an end goal.
However, we are interested in characterizing their variability and other properties as they will provide a set of initial conditions when AFL is restarted. In a fuzzing run on a single CPU core, we produce $936$ unique code paths used to train initial GAN and LSTM models. Random seed generation is performed by drawing random bytes from /dev/urandom.
For each method, we generate $200$ samples, reinitialize AFL on a single CPU with only the seed files of one method and run for an additional $72$ hours to measure the impact on code path discovery. 
Both LSTM and GAN models sightly out-perform random sampling for AFL reinitialization. 
We summarize the resulting mean time to discover new code paths in Table~\ref{tab:three}.

Each seed file produces a program trace when supplied as an input to ethkey.
Code paths that have different lengths will differ in at least one basic block or branch instruction.
The unique code path length is fast to compute but only provides a lower bound on the number of unique code paths exercised by the testing framework, across fuzzing runs using AFL. 
Two code paths with the same length can result from unique traces, thus detailed evaluation is needed to determine the true uniqueness of a code path from seed file execution. 





\begin{table}%
	\centering
	\begin{tabular}{ccccc}
		\hline
		Class C & $|C|$  & $L(C)$   & Sec/Path 	& Relative Rate  \\ \hline
		Urandom &	1231 &	0.9017	 & 214.478	&  1.00    \\ \hline
		LSTM &	1251 &		0.8984  &		197.130	& 1.08 \\ \hline
		GAN &	1240 &		0.8694	&		191.893	& 1.11 \\ \hline \\
	\end{tabular}
	\caption{{\bf Initial Run:} $|C|$ is the number of seeds generated after reinitializing AFL. We observe GAN or LSTM allows discovery of novel code paths at a quicker rate than restarting AFL using a random sampling of bytes.
$L(C)$ is the number of unique code path lengths $l(c_i)$ associated to input files $c_i$ in the set $C$ of LSTM, GAN, or uniform Random reinitialization.} \label{tab:three}%
	
\end{table}%

\begin{table}
  \centering
\begin{tabular}{cccccc }
\hline
Class $C$ & $|C|$ & $L(C)$ & \% Unique &  $\mu(L(C))$ & $\sigma(L(C))$ \\ \hline

AFL seed & 38384 & 31212 & 0.813 & 26.968M & 33.958M \\ \hline
Rand seed & 19824 & 485 & 0.024 & 2.602M & 724.674K \\ \hline
LSTM seed & 20000 & 1921 & 0.096 &  2.596M & 8.687K \\ \hline
GAN seed & 20000 &  119 & 0.006 & 2.593M  & 1.841K \\ \hline \\

\end{tabular}
\caption{{\bf Synthetic Seed Files:} Random sampling, LSTM, and GAN can be used to produce synthetic seed files for AFL.  We compute statistics on the synthetic files from the $3$ strategies.
  The synthetic files are not themselves deep or varied code paths, but can be used to reinitilize AFL.
}

\label{tab:one}%
\end{table}%

{\bf Large Experiment:}  To demonstrate the scalability of this augmentation strategy, we performed an extended run of AFL on $200$ CPU cores for $72$ hours.  Each core in the AFL run stopped finding seed files after the first 10 to 12 hours of fuzzing and accumulated a total of 39,185 seed files across 49 workers. 
All seed files produced within a given node are known to be unique, due to AFL's internal book keeping mechanism. 
However, seed files whose content are different across nodes, could in principle exercise the same code path.  By measuring the length of each program trace (code path), we can compute
a lower bound on the number of unique paths discovered by only counting paths that
have a unique length.
After removing identical seed files from across the nodes, and seed files that resulted in the same code path length, we estimate 802 of the initial files were duplicates from the independent worker nodes.  Removing those duplicates resulted in
a total of 38,384 unique files.  

We then trained GAN and LSTM networks on the total corpus of unique
seed files and generated approximately 20,000 samples from each method,
respectively, to use as synthetic seed files in order to reinitialize AFL.  GAN took approximately 30 minutes to train and generate synthetic seed files,
while LSTM took 14 hours to do so.

In Table~\ref{tab:one} we summarize the mean and variance of the length of program traces (i.e., code paths)
associated with the seed files from native AFL and from the synthetic generation methods for this larger experiment.
The synthetic seed files, when provided as inputs to the program under test, do not cause deep paths to be explored, compared to AFL.  So, we cannot simply {\em replace} AFL with a generative model.  Instead, we seek to combine generative models with AFL to {\em boost} its performance.
We see from this data that, in fact, the seed files generated by LSTM and GAN are not representative of the distribution $S$ in terms of the mean and variance of code paths generated.
This reinforces the need to use $S_G$ and $S_L$ as an augmentation strategy rather than a direct replacement of AFL seeds.

Next, we performed $24$
hours of fuzzing with GAN, LSTM, and a random reinitilialzation strategy using a random sampling of bytes from the initial seed files (i.e., performing no learning on the seed files). 
Table
~\ref{tab:five} summarizes our results.
All three strategies allowed for additional seed files to be
generated.  The GAN-based approach produced
seed files 14.23\% quicker than the random approach and 60.72\% faster than using LSTM.
We do lose 30 minutes of training time for GAN that could otherwise
be used for fuzzing using the random sampling method; discounting by this amount of time reduces
the code path rate to an 11.85\% improvement.
However, we are most interested in unique code paths.
GAN found the greatest number of seed files whose associated code paths
had lengths not found in the initial fuzzing run, outperforming the random control approach by 6.16\%.  
The average code path length discovered by GAN was 13.84\% longer than the random control,
so GAN is capable of exercising deeper paths in the program.
The LSTM model
underperformed both GAN and random sampling and took a substantially
longer time (14 hours) to train.

\begin{table}%
\centering
\begin{tabular}{cccccccc}
\hline
Class C & |C| & L(C) &  Novel  & L(C)  Rate     &  Novel Rate &  $\mu(L(C))$ & $\sigma(L(C))$ \\ \hline
Rand &   780	 &	778 &  682 &     1.000     & 1.000  &  25.373M &  3.339M \\ \hline
LSTM &  555 &	555 &   481 &    0.713       &  0.7053 &  26.541M & 3.385M \\ \hline
GAN &	891	 &	837 &    724   & 1.0758  &   1.0616 & 28.885M   &   3.456M \\ \hline \\
\end{tabular}
\caption{{\bf Sustained Run}:  With 38,000 training seed file samples, we compare the seed files generated after reinitialization from random, GAN, and LSTM generation methods.  L(C) Rate is the speedup over the random strategy of discovery of code paths with unique length per second.  Novel Rate is the speedup over random for unique lengths not found in the training set.
  \\}\label{tab:five}%
\end{table}%

\section{Conclusions}

In this work, we explored the utility of augmenting random mutation testing with deep neural models.  Natively AFL, combines file mutation strategies from Genetic Algorithms with program instrumentation via the use of compiler plugins.  We observed the most improvement in AFL's performance when we restart a fuzzing run mid-course, using novel seed files built from a GAN model.  Though the synthetic seed file statistics on average had similar path length, the GAN out-performed reinitialization from a random or LSTM strategy when restarting the fuzzing system.  The LSTM model was deficient in both training time and code path discovery time.  Both approaches used no manual analysis or information about file formats for the program under test.  The GAN and random strategies both improve the performance of AFL, even though the internal state of the program is never directly exposed.

Future work of interest includes experimentation on additional targets, including the DARPA Cyber Grand Challenge problems, open source OS network services, bytecode interpreters, and other system applications and programs where input data is easily generated.  We also plan to explore exposing the internal state of the program under test in order to define a reward function for reinforcement learning.  We envision this internal state could be exposed by: 1) the instrumentation AFL adds to programs via its GCC compiler plugins, 2) using Intel's PIN tool to output the length of each code path or summary information about a given trace 3) recording program traces using a replay framework such as Mozilla's rr tool in order to collect additional descriptive statistics.

\subsubsection*{Acknowledgements}


The authors would like to thank Court Corley, Nathan Hodas, and Sam Winters for useful discussions. The research described in this paper is part of the Deep Science Initiative at Pacific Northwest National Laboratory. It was conducted under the Laboratory Directed Research and Development Program at PNNL, a multi-program national laboratory operated by Battelle for the U.S. Department of Energy.

\clearpage
\medskip
\bibliography{acmsmall-sample-bibfile}
\end{document}